\pdfoutput=1

\documentclass[conference]{IEEEtran}
\IEEEoverridecommandlockouts
\usepackage{cite}
\usepackage{amsmath,amssymb,amsfonts,amsthm}
\usepackage{algorithmic}
\usepackage{graphicx}
\usepackage{xcolor}
\usepackage{bm,bbm}
\usepackage{booktabs}
\usepackage{color}
\usepackage{latexsym}
\usepackage{overpic}
\usepackage{cite}
\usepackage{url}
\usepackage{newtxtext,newtxmath}
\usepackage{hyperref}
\newtheorem{definition}{Definition}
\newtheorem{theorem}{Theorem}
\newtheorem{lemma}{Lemma}
\newcommand{\argmax}{\mathop{\rm arg\,max}\limits}

\newcommand{\pYX}{p_{\protect\scalebox{0.5}{Y|X}}}

\begin{document}
\title{Kernel Selection for Modal Linear Regression:\\Optimal Kernel and IRLS Algorithm\thanks{%
This paper will be published in the proceedings of the 18th IEEE International Conference on Machine Learning and Applications - ICMLA 2019.}\thanks{%
\copyright 2020 IEEE.  Personal use of this material is permitted.  Permission from IEEE must be obtained for all other uses, in any current or future media, including reprinting/republishing this material for advertising or promotional purposes, creating new collective works, for resale or redistribution to servers or lists, or reuse of any copyrighted component of this work in other works.}}
\author{
\IEEEauthorblockN{Ryoya Yamasaki}
\IEEEauthorblockA{\textit{Dept.~of Systems Science} \\
\textit{Grad.~Sch.~Info., Kyoto University}\\
Kyoto, Japan\\
yamasaki@sys.i.kyoto-u.ac.jp}
\and
\IEEEauthorblockN{Toshiyuki Tanaka}
\IEEEauthorblockA{\textit{Dept.~of Systems Science} \\
\textit{Grad.~Sch.~Info., Kyoto University}\\
Kyoto, Japan\\
tt@i.kyoto-u.ac.jp}
}
\maketitle
\begin{abstract}
Modal linear regression (MLR) is a method for obtaining a conditional mode predictor as a linear model.
We study kernel selection for MLR from two perspectives: ``which kernel achieves smaller error?'' and ``which kernel is computationally efficient?''.
First, we show that a Biweight kernel is optimal in the sense of minimizing an asymptotic mean squared error of a resulting MLR parameter.
This result is derived from our refined analysis of an asymptotic statistical behavior of MLR.
Secondly, we provide a kernel class for which iteratively reweighted least-squares algorithm (IRLS) is guaranteed to converge, and especially prove that IRLS with an Epanechnikov kernel terminates in a finite number of iterations.
Simulation studies empirically verified that using a Biweight kernel provides good estimation accuracy and that using an Epanechnikov kernel is computationally efficient.
Our results improve MLR of which existing studies often stick to a Gaussian kernel and modal EM algorithm specialized for it,
by providing guidelines of kernel selection.
\end{abstract}
\begin{IEEEkeywords}
Modal linear regression, 
Asymptotic normality, Optimal kernel, Biweight kernel, 
IRLS, Convergence, Epanechnikov kernel, 
Modal EM, Gaussian kernel
\end{IEEEkeywords}
\section{Introduction}\label{section:sec1}
Modal linear regression (MLR)~\cite{lee1989mode, lee1993quadratic, kemp2012regression, yao2014new} aims to obtain a conditional mode predictor consisting of a global maximizer of a conditional probability density function (PDF) of dependent variables conditioned on independent variables by using a linear model.
Besides a nice interpretability of a conditional mode, 
the MLR has an advantage that resulting parameter and curve estimators are consistent even for heteroscedastic or asymmetric conditional PDFs,
as compared with the robust M-type estimators which are not consistent~\cite{baldauf2012use}.
The consistency of the MLR even for data
with skewed conditional distributions
makes the MLR a promising approach to analyzing them; 
refer, for example, to an application to cognitive impairment prediction~\cite{NIPS20176743} and analysis of economic data~\cite{feng2017statistical, tian2017fitting}.
Thus studies of the MLR and related areas are currently ongoing from various viewpoints~\cite{tian2017fitting, sando2018information, ohta2018quantile}.

The MLR is a kernel-based semiparametric regression method, 
and the user has a freedom in selecting a kernel used in the MLR. 
However, many researches~\cite{zhao2014robust, yao2014new, Salah2017, sando2018information, LI201815} of the MLR use a Gaussian kernel, 
presumably because the modal EM algorithm (MEM)~\cite{li2007nonparametric, yao2014new}, 
a standard parameter estimation method for the MLR, 
does not provide an explicit parameter update formula
if a non-Gaussian kernel is used 
(see Section~\ref{section:sec2}).
In this paper, we study the problem of kernel selection in the MLR
from two perspectives: ``which kernel achieves smaller error?''
and ``which kernel is computationally efficient?''
in order to pave the way for use of a non-Gaussian kernel in the MLR. 

As far as the authors' knowledge,
relationship between the kernel used in the MLR
and the estimation error of the MLR has not been discussed at all. 
In Section~\ref{section:sec3} of this paper, we refine an existing theorem of asymptotic normality of the MLR parameter and derive the asymptotic mean squared error (AMSE) of it.
Then, adopting the AMSE as a criterion of an estimation error, we investigate the question~``which kernel achieves smaller error?''.
As a consequence, we find that a Biweight kernel is optimal among non-negative kernels for the MLR in the sense of minimizing the AMSE.

An objective function appearing in the MLR may have
multiple peaks and/or broad plateaux, which would make 
the optimization involved in the parameter estimation difficult. 
Accordingly, the parameter estimation in the MLR often needs a multi-start technique, i.e., repeating many times from different initial estimates.
So, it is plausible to have an efficient definite algorithm with convergence guarantees. 
In order to use non-Gaussian kernels in the MLR, 
parameter estimation methods other than the MEM are preferable, 
because the non-Gaussian MEM needs to use an additional optimization algorithm in its inner loop and may be inefficient.
In Section~\ref{section:sec4} of this paper, 
we consider using the iteratively reweighted least-squares algorithm (IRLS) for the MLR and show that the IRLS has a convergence guarantee for a broad class of kernels.
Also, as a partial answer to the question ``which kernel is computationally efficient?'', we prove a theorem stating that the IRLS exactly terminates in a finite number of iterations if an Epanechnikov kernel is used.

In Section~\ref{section:sec5}, we show results of numerical experiments
of the MLR with several kernels on simulation data.
There, we confirmed that the mean squared error (MSE) when using a Biweight kernel became the smallest as the sample size went larger, 
and that the calculation efficiency of the IRLS with an Epanechnikov kernel was much higher than that using other kernels.
Finally, conclusions are summarized in Section~\ref{section:sec6}.
The proofs of the theoretical results are in Appendices.

\section{Existing researches on MLR and MEM}\label{section:sec2}
Let $\mathcal{X}\subset\mathbb{R}^p$ be the input space 
and $\mathcal{Y}\subset\mathbb{R}$ be the output space.
Let $(\bm{X},Y)$ be a pair of random variables taking values in 
$\mathcal{X}\times\mathcal{Y}$ following a certain joint distribution. 
We assume that the conditional distribution with the PDF $\pYX(\cdot|\bm{x})$
of $Y$ conditioned on $\bm{X}=\bm{x}\in\mathcal{X}$ 
is such that for a certain function $\tilde{m}$ 
the residual $\epsilon=Y-\tilde{m}(\bm{x})$ 
is distributed according to a distribution whose mode is at the origin. 
It then follows that $\text{arg max}_y\pYX(y|\bm{x})=\tilde{m}(\bm{x})$ holds
for any $\bm{x}\in\mathcal{X}$. 
In the MLR, it is further assumed that $\tilde{m}$ is a linear function,
\begin{align}
\label{eq:CMF}
\tilde{m}(\bm{x})=\tilde{\bm{\theta}}^{\top}\bm{x},
\end{align}
where $\tilde{\bm{\theta}}\in\mathbb{R}^p$ is an underlying MLR parameter.

Suppose that a sample set $\{(\bm{x}_i,y_i)\in\mathcal{X}\times\mathcal{Y}\}_{i=1}^n$
consists of $n$ independent and identically-distributed (iid) 
samples from the joint distribution of $(\bm{X},Y)$. 
To estimate the underlying parameter $\tilde{\bm{\theta}}$ in~\eqref{eq:CMF} on the basis of the sample set, 
the MLR maximizes the kernel-based objective function
\begin{align}
\label{eq:OBJ}
O(\bm{\theta})
=\frac{1}{n}\sum_{i=1}^n K_h\left(y_i-\bm{\theta}^{\top}\bm{x}_i\right),
\end{align}
where $K_h$ is the kernel function such that $K_h(\cdot)=h^{-1}K(\cdot/h)$ for some function $K$, which is also called a kernel function, 
and  where $h>0$ is a scale parameter (bandwidth) of $K_h$.
Also, the estimator of the MLR parameter, defined as the maximizer of \eqref{eq:OBJ}, is denoted as $\hat{\bm{\theta}}_n$.
\cite{kemp2012regression} and \cite{yao2014new} have provided fundamental analysis of asymptotic statistical behaviors of the MLR, 
such as consistency and asymptotic normality of the estimator $\hat{\bm{\theta}}_n$.

\cite{yao2014new} has tackled the optimization of~\eqref{eq:OBJ} by applying the idea of the MEM~\cite{li2007nonparametric}.
It alternately iterates the following two steps from a given initial parameter estimate $\bm{\theta}_0\in\mathbb{R}^p$:
\begin{align}
\label{eq:MEM-Estep}
&\text{E-step:}\quad p_{hit}
=\frac{K_h\left(y_i-\bm{\theta}_t^{\top}\bm{x}_i\right)}
{\sum_{j=1}^n K_h\left(y_j-\bm{\theta}_t^{\top}\bm{x}_j\right)},\quad i=1,\ldots,n,\\
\label{eq:MEM-Mstep}
&\text{M-step:}\quad\bm{\theta}_{t+1}
=\argmax_{\bm{\theta}\in\mathbb{R}^p}
\sum_{i=1}^n p_{hit}\log K_h\left(y_i-\bm{\theta}^{\top}\bm{x}_i\right).
\end{align}
When $K_h$ is a Gaussian kernel, the M-step \eqref{eq:MEM-Mstep} admits an analytic expression, 
\begin{align}
\label{eq:MEM-Gau}
\bm{\theta}_{t+1}=(\mathrm{X}^{\top}\mathrm{K}_{ht}\mathrm{X})^{-1}
\mathrm{X}^{\top}\mathrm{K}_{ht}\bm{y},
\end{align}
where $\mathrm{X}=(\bm{x}_1,\ldots,\bm{x}_n)^{\top}$,
$\bm{y}=(y_1,\ldots,y_n)^{\top}$, 
and $\mathrm{K}_{ht}$ is an $n\times n$ diagonal matrix with diagonal elements $p_{hit}\propto K_h(y_i-\bm{\theta}_t^{\top}\bm{x}_i)$. 
The MEM~\cite{li2007nonparametric, yao2014new}, as shown above, is like the well-known EM algorithm.
It has been shown that, irrespective of the kernel used, the objective function sequence $\{O(\bm{\theta}_t)\}$ is non-decreasing and converges, 
as long as the M-step~\eqref{eq:MEM-Mstep} is executed exactly. 

Also, several studies have combined the MLR with regularization, variable selection, and other techniques~\cite{yu2012bayesian, feng2017statistical, zhao2014robust, Salah2017, LI201815}.
The MEM has been used in most of them,
and Gaussian kernels have been used almost exclusively. 
For instance, \cite{feng2017statistical} has studied relationships between the MLR using a Gaussian kernel and a correntropy-based regression~\cite{liu2007correntropy, feng2015learning}.
Also, when the bandwidth $h$ is quite large, 
it is known that a Gaussian kernel $K_h$ is approximated as $\exp(-(u/h)^2/2)\approx 1-(u/h)^2/2$.
Thus, \cite{kemp2012regression} has discussed that the MLR with a Gaussian kernel approaches the ordinary least-squares (OLS) as $h\to\infty$.

\section{MSE-based Asymptotics and Optimal Kernel}\label{section:sec3}
\subsection{Asymptotic Normality of MLR Parameter}
We discuss the optimal kernel minimizing an estimation error for the MLR in the next subsection, 
on the basis of the asymptotic statistical theory.
For this purpose, we review in this section the analysis on asymptotic behaviors of the MLR.
In the standard asymptotic analysis,
in order to have consistency of the estimator $\hat{\bm{\theta}}_n$, 
the bandwidth $h$ of the kernel $K_h$ should be made smaller 
as the sample size $n$ tends to be larger, but not too fast
in order to prevent $O(\bm{\theta})$ from becoming rough.
A bandwidth sequence $\{h_n\}$ specifies
how one sets the bandwidth $h$ according to the sample size $n$.
It, together with the kernel used, determines
the asymptotic behaviors of the MLR. 

Among the existing studies which discuss the asymptotics the most strict is that by~\cite{kemp2012regression}.
The following theorem is a refinement of their result. 
\begin{theorem}
\label{theorem:Theorem1}
Let $K$ be a bounded function with 
the bounded second-order moment and 
the first three bounded derivatives $K^{(j)}$, $j=1,2,3$, 
of which $K^{(j)}$, $j=1,2$ are square-integrable, satisfying
\begin{align}
\label{eq:KerAsm}
\begin{split}
&\lim_{|u|\to\infty}u^3K(u)=0,\ 
\int K(u)\,du=1,\ 
\int uK(u)\,du=0,\\
&\int u^2K(u)\,du=U\ (0<|U|<\infty),\ 
\int \{K^{(1)}(u)\}^2\,du=V,
\end{split}
\end{align}
and let $\{h_n\}$ be a sequence of positive constants satisfying 
\begin{align}
\label{eq:order}
h_n\to0,\quad
nh_n^5/\log(n)\to\infty,
\end{align}
as $n\to\infty$. 
Then, under the regularity conditions in Appendix~\hyperlink{A}{A}, 
the MLR parameter estimate sequence $\{\hat{\bm{\theta}}_n\}$
obtained with the bandwidth sequence $\{h_n\}$ asymptotically has a normal distribution: 
\begin{align}
\label{eq:normal}
(nh_n^3)^{1/2}\left(\hat{\bm{\theta}}_n-\tilde{\bm{\theta}}-h_n^2U\tilde{\mathrm{A}}^{-1}\tilde{\bm{b}}/2\right)
\overset{D}{\to}\mathcal{N}
\left(\bm{0}, V\tilde{\mathrm{A}}^{-1}\tilde{\mathrm{C}}\tilde{\mathrm{A}}^{-1}\right),
\end{align}
where $\tilde{\mathrm{A}}=\mathrm{E}[\pYX^{(2)}(\tilde{\bm{\theta}}^\top\bm{X}|\bm{X})\bm{X}\bm{X}^\top]$, $\tilde{\bm{b}}=\mathrm{E}[\pYX^{(3)}(\tilde{\bm{\theta}}^\top\bm{X}|\bm{X})\bm{X}]$, and 
$\tilde{\mathrm{C}}=\mathrm{E}[\pYX(\tilde{\bm{\theta}}^\top\bm{X}|\bm{X})\bm{X}\bm{X}^\top]$.
\end{theorem}

The outline of the proof is in Appendix~\hyperlink{A}{A}.

In~\cite{kemp2012regression},
the scaling $h_n=o(n^{-1/7})$ was assumed, and accordingly, the asymptotic bias,
which is proportional to $(nh_n^3)^{1/2}h_n^2=o(n^0)\to0$ as $n\to\infty$,
was treated as zero. 
Theorem~\ref{theorem:Theorem1} provides the explicit expression of the asymptotic bias by relaxing this assumption. 
This modification is crucial in the following discussion.

\subsection{Optimal Kernel for MLR}
\begin{table*}[t]
\centering
\caption{Value of the AMSE criterion $U^{6/7}V^{4/7}$ for each kernel, and whether it is QM or not.}
\label{tb:OPT}
\begin{small}\begin{sc}
\begin{tabular}{cc|cccc|c}
\toprule
Kernel & $K(u)$ & $U$ & $V$ & $U^{6/7}V^{4/7}$ & $\cdot/0.2916$ & QM at $u'$?\\
\midrule
Biweight& $\frac{15}{16}\{(1-u^2)_+\}^2$& 
$\frac{1}{7}$ & $\frac{15}{7}$ & {\bf0.2916}&{\bf 1.0000} &$\surd$\\
Triweight& $\frac{35}{32}\{(1-u^2)_+\}^3$& 
$\frac{1}{9}$ & $\frac{35}{11}$ & 0.2947&1.0105&$\surd$\\
Tricube& $\frac{70}{81}\{(1-|u|^3)_+\}^3$& 
$\frac{35}{243}$ & $\frac{420}{187}$ & 0.3016&1.0345&$\times$\\
Cosine& $\frac{\pi}{4}\cos(\frac{\pi u}{2})\mathbbm{1}_{|u|\le1}$ & 
$1-\frac{8}{\pi^2}$ & $\frac{\pi^4}{64}$ & 0.3054&1.0475&$\surd$\\
Epanechnikov& $\frac{3}{4}(1-u^2)_+$& 
$\frac{1}{5}$ & $\frac{3}{2}$ & 0.3173&1.0883&$\surd$\\
Triangle& $(1-|u|)_+$& 
$\frac{1}{6}$ & 2 & 0.3199&1.0971&excl.~$u'=0$\\
Gaussian& $\frac{1}{(2\pi)^{1/2}}e^{-u^2/2}$ & 
1 & $\frac{1}{4\pi^{1/2}}$ & 0.3265&1.1198&$\surd$\\
Logistic& $\frac{1}{e^u+2+e^{-u}}$ & 
$\frac{\pi^2}{3}$ & $\frac{1}{30}$ & 0.3974&1.3629&$\surd$\\
Laplace& $\frac{1}{2}e^{-|u|}$ & 
2 & $\frac{1}{4}$ & 0.8203&2.8133&excl.~$u'=0$\\
Sech& $\frac{1}{2}{\rm sech}(\frac{\pi u}{2})$ & 
1 & $\frac{\pi^2}{12}$ & 0.8943&3.0671&$\surd$\\
\bottomrule
\end{tabular}
\end{sc}\end{small}
\end{table*}
From Theorem~\ref{theorem:Theorem1},
keeping only the leading-order terms under $n\to\infty$, 
the bias and the variance of $\hat{\bm{\theta}}_n$
are given by $h_n^2 U \tilde{\mathrm{A}}^{-1}\tilde{\bm{b}}/2$
and $V \mathrm{tr}(\tilde{\mathrm{A}}^{-1}\tilde{\mathrm{C}}\tilde{\mathrm{A}}^{-1})/(nh_n^3)$,
respectively, where $\mathrm{tr}(\mathrm{A})$ denotes the trace of a matrix $\mathrm{A}$.
Accordingly, 
the AMSE of $\hat{\bm{\theta}}_n$ 
is given as a sum of the squared asymptotic bias
and the asymptotic variance, as 
\begin{align}
\label{eq:AMSE}
	\mathrm{E}[\|\hat{\bm{\theta}}_n-\tilde{\bm{\theta}}\|^2]
	\approx \frac{h_n^4U^2\|\tilde{\mathrm{A}}^{-1}\tilde{\bm{b}}\|^2}{4}
	+\frac{V\mathrm{tr}\left(\tilde{\mathrm{A}}^{-1}\tilde{\mathrm{C}}\tilde{\mathrm{A}}^{-1}\right)}{nh_n^3}.
\end{align}
The ratio of the squared-bias to the variance is $O(nh_n^7)$.
Which of the squared-bias and the variance is dominant
in the AMSE depends on the rate of decay
of the bandwidth sequence $\{h_n\}$ as $n\to\infty$:
If $\{h_n\}$ decays slowly, then it will be dominated by the squared-bias,
whereas if $\{h_n\}$ decays fast, then it will be dominated by the variance.
The fastest decay of the AMSE is achieved 
at the boundary between the bias-dominant and variance-dominant regimes, i.e., when $h_n=O(n^{-1/7})$. 

Moreover, on the basis of the stationary condition of the AMSE with respect to $h_n$, 
the asymptotic optimal bandwidth minimizing the AMSE becomes
\begin{align}
\label{eq:optH}
\tilde{h}_n=
\left[\frac{3V \mathrm{tr}\left(\tilde{\mathrm{A}}^{-1}\tilde{\mathrm{C}}\tilde{\mathrm{A}}^{-1}\right)}
{n U^2 \|\tilde{\mathrm{A}}^{-1}\tilde{\bm{b}}\|^2}\right]^{1/7},
\end{align}
and under this optimal bandwidth $\tilde{h}_n$ 
the leading-order term of the AMSE becomes proportional to
\begin{align}
\label{eq:OPTAMSE}
U^{6/7}V^{4/7}n^{-4/7}
\|\tilde{\mathrm{A}}^{-1}\tilde{\bm{b}}\|^{6/7}
\left\{\mathrm{tr}\left(\tilde{\mathrm{A}}^{-1}\tilde{\mathrm{C}}\tilde{\mathrm{A}}^{-1}\right)\right\}^{4/7}.
\end{align}
This relationship implies that the optimal convergence rate of the AMSE is $n^{-4/7}$, 
but we would like to focus here on the dependence not on $n$ but rather on the kernel-dependent quantities $U$ and $V$: 
indeed, quantities with tilde in~\eqref{eq:OPTAMSE} are determined from the underlying data distribution, 
whereas $U$ and $V$ are completely determined by the kernel $K$.
Thus, the optimal kernel should be obtained by minimizing $U^{6/7}V^{4/7}$ (called the AMSE criterion in this paper). 
It is a variational problem, whose solution is given by:
\begin{theorem}
\label{theorem:Theorem2}
A Biweight kernel $K(u)=(15/16)\{(1-u^2)_+\}^2$ minimizes the AMSE criterion $U^{6/7}V^{4/7}$ 
among non-negative kernels.
\end{theorem}

This result is derived from the theory developed in~\cite{granovsky1991optimizing} on the optimal kernel for mode estimation, where
the same variational problem appears. 
We give the outline of the proof in Appendix~\hyperlink{B}{B}.

In order to check the degree of goodness of a Biweight kernel
in comparison with other kernels, 
we have calculated the criterion $U^{6/7}V^{4/7}$ for various commonly-used kernel functions, 
ignoring their differentiability.
The results are summarized in Table~\ref{tb:OPT}.
The AMSE is about 12 percent larger for a Gaussian kernel than for a Biweight kernel.
An empirical observation is 
that truncated kernels such as a Triweight and an Epanechnikov in addition to a Biweight are better in terms of the AMSE criterion than a Gaussian kernel.
Also, the AMSE criterion becomes larger for heavy-tailed kernels including a Laplace and a Sech, and 
we may conclude that heavy-tailed kernels are not good for use in the MLR.

\section{IRLS and its Convergence}\label{section:sec4}
\subsection{Construction of IRLS for MLR from MM Algorithm}
\begin{figure*}[t]
\centering
\begin{minipage}{0.245\hsize}\centering
\begin{overpic}[clip, width=4.5cm, bb=0 0 675 550]{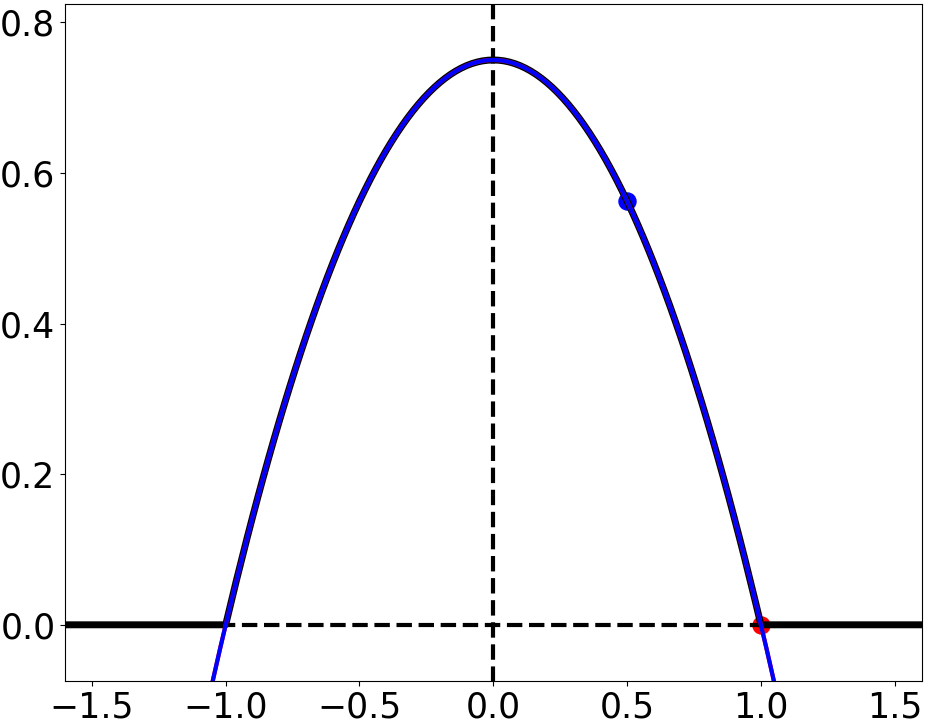}
\put(12,65){\sf (a)}\end{overpic}\end{minipage}
\begin{minipage}{0.245\hsize}\centering
\begin{overpic}[clip, width=4.5cm, bb=0 0 675 550]{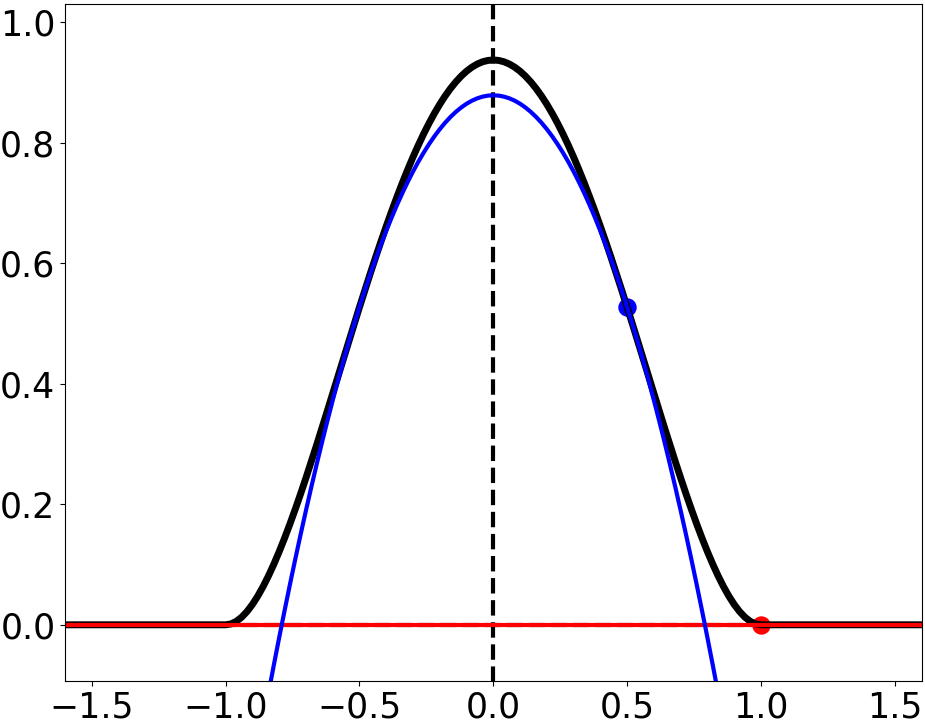}
\put(12,65){\sf (b)}\end{overpic}\end{minipage}
\begin{minipage}{0.245\hsize}\centering
\begin{overpic}[clip, width=4.5cm, bb=0 0 675 550]{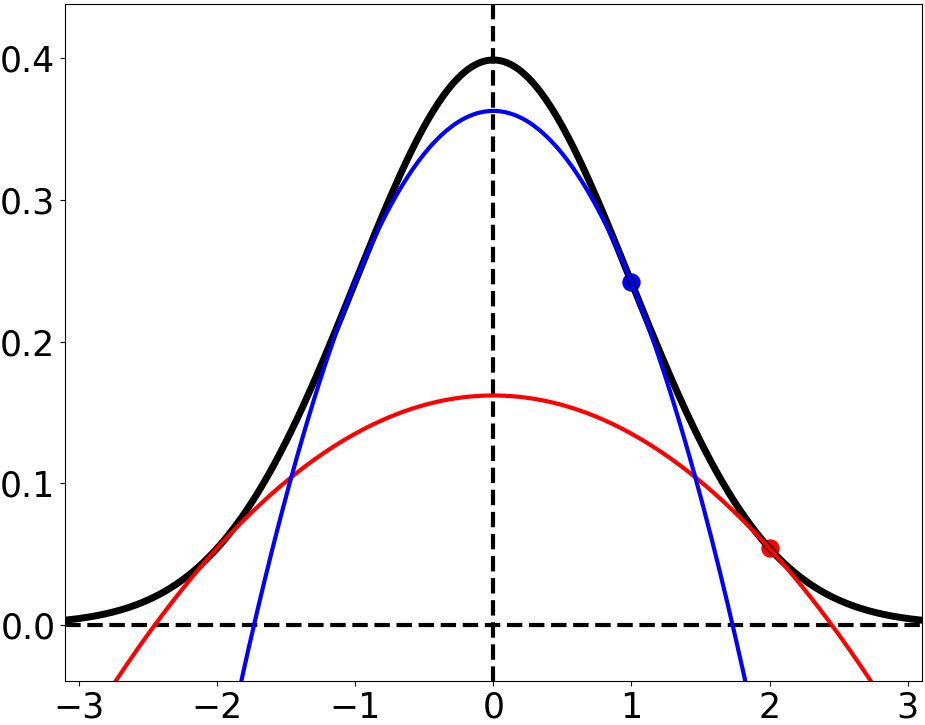}
\put(12,65){\sf (c)}\end{overpic}\end{minipage}
\begin{minipage}{0.245\hsize}\centering
\begin{overpic}[clip, width=4.5cm, bb=0 0 675 550]{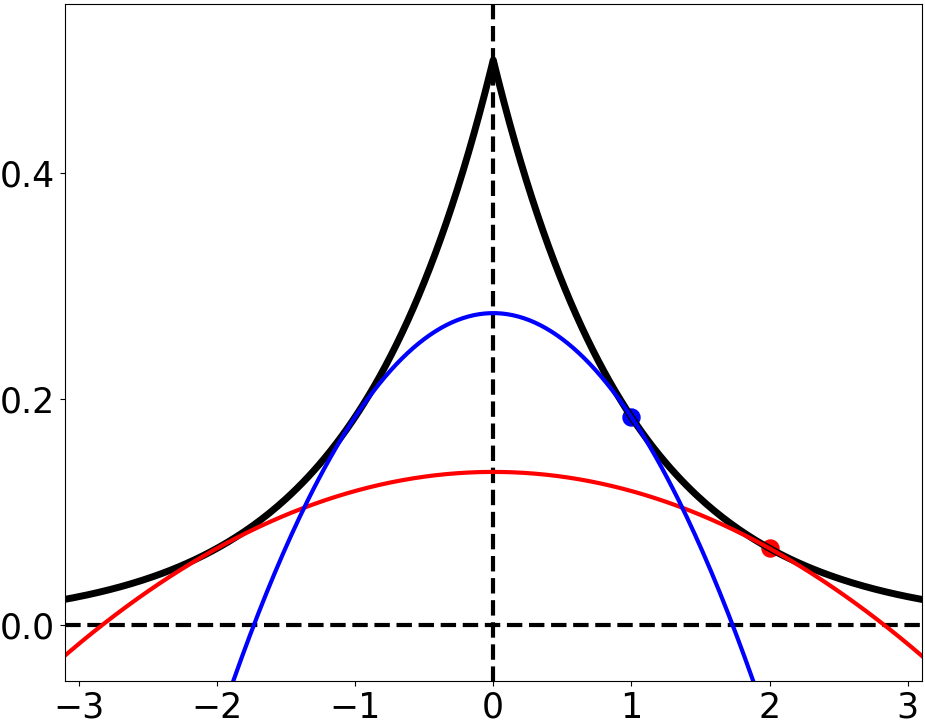}
\put(12,65){\sf (d)}\end{overpic}\end{minipage}
\caption{
The QM kernel $K_h$ and the best quadratic minorizer $G_h(\cdot|u')$ of $K_h$ at $u'$.
The black rigid line represents $K_h$, and the red and blue lines represent $G_h(\cdot|u')$ at different points of red and blue, respectively.
(a)-(d) are for the Epanechnikov, Biweight, Gaussian, and Laplace kernels, respectively.
Note that the two red and blue lines of (a) overlap.}
\label{img:img-ker}
\end{figure*}

As mentioned in Section~\ref{section:sec2},
a plausible property of the MEM is 
that the objective function sequence $\{O(\bm{\theta}_t)\}$ given by the MEM converges no matter what kernel is used. 
It would presumably be the main reason
as to why the MEM has frequently been used. 
However, the maximization on the right-hand side of~\eqref{eq:MEM-Mstep} can explicitly be solved only when a Gaussian kernel is used.
If using a non-Gaussian kernel, 
the M-step requires use of an additional optimization as an inner loop, such as the conventional gradient method.
This may reduce computational efficiency and even suffer from convergence problem of the inner loop.
In contrast, we consider in this section an IRLS-based parameter estimation method, 
which provides an explicit parameter update formula even when a non-Gaussian kernel is used.
Although several existing works~\cite{kemp2012regression, feng2017statistical, NIPS20176743} using the IRLS for the MLR have not discussed its convergence properties, 
we clarify in this paper its convergence properties on the basis of reformulation of the IRLS based on the minorize-maximize (MM) algorithm~\cite{lange2016mm}.

We first assume that the kernel $K_h$ is symmetric about the origin.
Hence, it is represented as $K_h(u)=\bar{k}_h((u/h)^2)$ with a function $\bar{k}_h$, 
called a profile of $K_h$, where the term `profile' is defined as follows.
\begin{definition}[profile]
A function $\bar{f}$ defined on $\mathbb{R}_{\ge0}$ is called a \emph{profile} (with coefficient $\gamma$) of a function $f$ on $\mathbb{R}$, 
if $f$ can be represented as $f(u)=\bar{f}(\gamma u^2)$ 
with a constant $\gamma>0$.
\end{definition}

We further assume that the kernel function $K_h$ is quadratic minorizable as defined below.
To that end, we first provide the definition of the minorizer.
\begin{definition}[minorizer]
We say that a function $h(\cdot|u')$ is a minorizer of a function $f(\cdot)$ at $u'$, 
if it satisfies
\begin{align}
\label{eq:Mino}
h(u'|u')=f(u');\quad
h(u|u')\le f(u),\quad \forall u.
\end{align}
\end{definition}

Following this definition, 
notions concerning the quadratic minorizability are given as follows:
\begin{definition}[quadratic minorizability]
\label{def:QM}
A \emph{quadratic minorizer} of $f(\cdot)$ at $u'$ is defined
as a minorizer of $f(\cdot)$ at $u'$ that is quadratic or constant. 
We call $f(\cdot)$ \emph{quadratically minorizable} (QM) at $u'$
if it has a quadratic minorizer at $u'$. 
Furthermore, when $f(\cdot)$ is QM at $u'$,
among quadratic minorizers of $f(\cdot)$ at $u'$,
the one whose vertex has the largest value is called \emph{the best quadratic minorizer} of $f(\cdot)$ at $u'$.
\end{definition}

Then, one sufficient condition that a function has a quadratic minorizer is provided as follows (see~\cite{de2009sharp} for the details).
\begin{lemma}
\label{lemma:Lemma1}
If a function $f$ has a continuous, convex, and non-increasing profile $\bar{f}$ with coefficient $\gamma>0$,  
$f$ is quadratically minorizable at any non-zero point. 
Also, the best quadratic minorizer $h(\cdot|u')$ of $f(\cdot)$ at $u'\neq0$ is 
\begin{align}
\label{eq:MQM}
h(u|u')
=g(u')u^2
+\left(f(u')-g(u'){u'}^2\right),
\end{align}
where $g(u')=\gamma \check{f}(\gamma u'^2)$ and where $\check{f}(u')$ is the minimum value among the subderivatives\footnote{%
A subderivative of a convex function $f$ at a point $u'$ is a real number $c$ such that $f(u)-f(u')\ge c(u-u')$, and thus it takes values in an interval between $\lim_{u\downarrow u'}\tfrac{f(u)-f(u')}{u-u'}$ and $\lim_{u\uparrow u'}\tfrac{f(u)-f(u')}{u-u'}$.} 
of $\bar{f}$ at $u'$.
\end{lemma}

From the viewpoints mentioned thus far, we introduce the following class of kernels for the MLR and IRLS.
\begin{definition}[QM kernel]
\label{def:QM-K}
A kernel function $K_h$ is called a \emph{QM kernel} 
if $K_h$ is non-negative, continuous, and normalized and its profile $\bar{k}_h$ is convex and non-increasing.
\end{definition}

Although the QM kernel is more restrictive than the `modal regression kernel' defined in \cite{feng2017statistical},
it includes a wide variety of practically-used kernel functions (see Table~\ref{tb:OPT} and Figure~\ref{img:img-ker}).
However, the QM kernel does not generally satisfy the conditions of Theorem~\ref{theorem:Theorem1}, mainly with regard to its differentiability.

For the QM kernel $K_h$, 
its best quadratic minorizer is represented as 
\begin{align}
\label{eq:K-M}
G_h(u|u')
=g_h(u')u^2+\left(K_h(u')-g_h(u')u'^2\right),
\end{align}
where $g_h(u')=\check{k}_h((u'/h)^2)/h^2$ when $u'\neq0$,
where $\check{k}_h$ is the minimal value of the subderivative of the profile $\bar{k}_h$.
However, for a kernel with a profile satisfying $\lim_{u\downarrow0}d\bar{k}_h(u)/du\neq-\infty$, 
$g_h(u')$ is ill-defined at $u'=0$ and one cannot construct a quadratic minorizer when $u'$ is exactly equal to 0.
For such a kernel, 
we assume that $y_i-\bm{\theta}_t^\top\bm{x}_i$, $i=1,\ldots,n$, does not become exactly zero for all $t$. 
We also note that $g_h$ is non-positive, since it is assumed that the profile of the QM kernel is non-increasing.
Then, given $\bm{\theta}_t$,
we can construct a quadratic minorizer of the objective function~\eqref{eq:OBJ} as 
\begin{align}
\label{eq:OBJ-M}
O_M(\bm{\theta}|\bm{\theta}_t)=\frac{1}{n}\sum_{i=1}^n
G_h\left(y_i-\bm{\theta}^{\top}\bm{x}_i
|y_i-\bm{\theta}_t^{\top}\bm{x}_i\right).
\end{align}
Moreover, on the basis of the stationary condition of the quadratic minorizer,
\begin{align}
\label{eq:OBJ-M-stat}
\frac{\partial}{\partial\bm{\theta}}O_M(\bm{\theta}|\bm{\theta}_t)=
-\frac{2}{n}\sum_{i=1}^n g_{hit}\cdot(y_i-\bm{\theta}^{\top}\bm{x}_i)\bm{x}_i=\bm{0},
\end{align}
where $g_{hit}=g_h(y_i-\bm{\theta}^{\top}_t\bm{x}_i)$, 
we construct the IRLS-based parameter update formula
\begin{align}
\label{eq:MM-IRLS}
\bm{\theta}_{t+1}=
\left(\sum_{i=1}^n g_{hit}\bm{x}_i\bm{x}_i^{\top}\right)^{-1}
\sum_{i=1}^n g_{hit}y_i\bm{x}_i,
\end{align}
or equivalently 
$\bm{\theta}_{t+1}=
(\mathrm{X}^{\top}\mathrm{G}_{h t}\mathrm{X})^{-1}\mathrm{X}^{\top}\mathrm{G}_{ht}\bm{y}$, 
where $\mathrm{G}_{ht}$ is an $n\times n$ diagonal matrix with diagonal elements $g_{hit}$.
When a Gaussian kernel is used, the IRLS~\eqref{eq:MM-IRLS} gives the same update formula as the MEM because of the relationship $g_{hit}\propto p_{hit}$.

\subsection{Convergence of IRLS for MLR}
\begin{table*}[t]
\centering
\caption{MSE and standard deviation of MSE (both are multiplied by $10^{2}$)}
\label{tb:SE}
\begin{small}\begin{sc}
\begin{tabular}{c|cccccccr}
\toprule
Kernel & $n=100$ & $n=200$ & $n=400$ & $n=800$ & $n=1600$ & $n=3200$ & $n=6400$\\
\midrule
Epanechnikov&
24.335$\pm$1.166&10.745$\pm$0.474&5.086$\pm$0.225&2.667$\pm$0.114&1.465$\pm$0.064&0.838$\pm$0.038&0.513$\pm$0.021\\
Biweight&
21.136$\pm$1.087&{\bf9.527$\pm$0.419}&{\bf4.574$\pm$0.192}&{\bf2.470$\pm$0.108}&{\bf1.358$\pm$0.059}&{\bf0.787$\pm$0.035}&{\bf0.449$\pm$0.019}\\
Gaussian&
{\bf20.947$\pm$1.056}&9.710$\pm$0.431&4.683$\pm$0.188&2.661$\pm$0.115&1.457$\pm$0.064&0.845$\pm$0.036&0.486$\pm$0.020\\
Laplace&
38.726$\pm$1.711&20.147$\pm$0.984&10.281$\pm$0.451&5.628$\pm$0.246&3.357$\pm$0.149&1.896$\pm$0.089&1.343$\pm$0.055\\
\bottomrule
\end{tabular}
\end{sc}\end{small}
\end{table*}

Since the IRLS is based on the MM algorithm, 
its ascent property and convergence can be proved from the general theory of the MM algorithm.
The following convergence result of the IRLS is a direct consequence of its construction.
\begin{theorem}
\label{theorem:Theorem3}
Let $K_h$ be a QM kernel.
Then, for the parameter estimate sequence $\{\bm{\theta}_t\}$ obtained via the IRLS, 
the objective function sequence $\{O(\bm{\theta}_t)\}$ is non-decreasing and converges.
\end{theorem}

This theorem implies that the IRLS is superior to the MEM in that the former allows for a variety of kernels to be practically used in the MLR
with convergence guarantee. 

We furthermore provide a stronger result 
stating that the IRLS with an Epanechnikov kernel
converges in a finite number of iterations.
The proof, detailed in Appendix~\hyperlink{C}{C},
is on the basis of the fact that the subderivative $g_{hit}$
takes only a finite number of values
when an Epanechnikov kernel is used. 
\begin{theorem}
\label{theorem:Theorem4}
Consider the parameter estimate sequence $\{\bm{\theta}_t\}$ obtained via the IRLS using an Epanechnikov kernel $K(u)=(3/4)(1-u^2)_+$.
Then, there exists a finite $\tau$
such that $O(\bm{\theta}_t)=O(\bm{\theta}_\tau)$ holds for all  $t\ge\tau$. 
Moreover, if the Hessian of $O(\bm{\theta})$ at $\bm{\theta}=\bm{\theta}_\tau$ is negative definite, $\bm{\theta}_t=\bm{\theta}_\tau$ holds for all $t\ge\tau$. 
\end{theorem}

\section{Simulation Studies}\label{section:sec5}
The objective of the numerical experiments described in this section
is \emph{neither} to compare the MLR with other regression methods
including OLS, least absolute deviation, Huber's robust regression,
and so on, as in~\cite{kemp2012regression, yao2014new},
\emph{nor} to try applying the MLR to real-world data.
Rather, the objective is to verify our theoretical results
on effects of the kernel function on the performance of the MLR. 

In reference to the experiment design in~\cite{yao2014new}, 
we generated datasets consisting of $n=100, 200, \ldots, 6400$ iid samples from the heteroscedastic distribution $Y=1+3X_2+(1+2X_2)\epsilon$ 
with $X_2\sim\mathcal{U}([0,1])$ (let $\bm{X}=(1,X_2)^\top$) 
and $\epsilon\sim0.5\mathcal{N}(-1,3^2)+0.5\mathcal{N}(1,0.3^2)$
being independent, 
where $\mathcal{U}(s)$ is a uniform distribution with support $s$. 
Thus, the underlying conditional mode becomes $\tilde{m}(\bm{X})=1+3X_2+(1+2X_2)\bar{m}\approx2+5X_2$ since $\bar{m}=\mathrm{Mode}(\epsilon)=0.9897\ldots\approx1$, 
and hence the underlying parameter $\tilde{\bm{\theta}}$ is
given by $\tilde{\bm{\theta}}=(\tilde{\theta}_1,\tilde{\theta}_2)^\top=(1+\bar{m},3+2\bar{m})^\top$.

We used the linear model $Y=\bm{\theta}^\top \bm{X}$, where $\bm{\theta}=(\theta_1, \theta_2)^\top$, and the IRLS~\eqref{eq:MM-IRLS} as a parameter estimation method\footnote{%
We also experimented the MEM using a conventional gradient method in its inner loop (M-step), but it often stopped in a plateau when using non-Gaussian kernels.}.
We adopted the multi-start method with 10 points randomly drawn from $\mathcal{U}([\tilde{\theta}_1-0.1,\tilde{\theta}_1+0.1]\times[\tilde{\theta}_2-0.1,\tilde{\theta}_2+0.1])$ 
being used for the initial parameter $\bm{\theta}_0$, 
in order to mitigate issues arising from the multimodality of an objective function.
We judged convergence by $\|\bm{\theta}_{t+1}-\bm{\theta}_t\|\le10^{-4}$.
We compared four different kernels: 
Epanechnikov $K_h(u)=(3/(4h))(1-(u/h)^2)_+$, Biweight $K_h(u)=(15/(16h))\{(1-(u/h)^2)_+\}^2$, Gaussian $K_h(u)=(2\pi h^2)^{-1/2}e^{-(u/h)^2/2}$, and Laplace $K_h(u)=(1/(2h))e^{-|u/h|}$.
It should be noted that the Laplace kernel is not QM at $0$, 
but in our experiments, 
we did not encounter a serious trouble caused by the update formula~\eqref{eq:MM-IRLS} becoming ill-defined as described in Section~\ref{section:sec4}.
For the bandwidth $h=h_n$ of the kernel function,
we used the empirical counterpart of the optimal bandwidth in~\eqref{eq:optH}, 
on the basis of the datasets and the underlying data distribution.
As an evaluation criterion of the MLR in our experiments, 
we adopted the MSE between $\hat{\bm{\theta}}_n$ and $\tilde{\bm{\theta}}$, 
which was calculated on the basis of 1000 trials.
We report the MSE and the standard deviation of the MSE in Table~\ref{tb:SE}.

When the sample size is the smallest, $n=100$, 
the best result was obtained when the Gaussian kernel was used.
However, increasing the sample size resulted in the smallest MSE when the Biweight kernel was used.
Also, the MSE with the Laplace kernel was the largest for every $n$.
These results are in good agreement with the results of the asymptotic analysis\footnote{%
Our analysis in Section~\ref{section:sec3} is based on the asymptotics concerning when the sample size $n$ is large enough. 
Readers may want to know the theory for the small sample case, but there is no valid result for kernel selection in such a case, as far as we are aware of.}
given in Theorem~\ref{theorem:Theorem2} and Table~\ref{tb:OPT}.

The ranking of the calculation time of the IRLS for each kernel did not change even if the sample size $n$ was varied.
Therefore, we report the calculation times with $n=6400$:
the IRLS converged in $5.9$, $11.5$, $16.0$, and $145.7$ seconds from an initial point on average for the Epanechnikov, Biweight, Gaussian, and Laplace kernels, respectively.
Thus, the computational efficiency with the Epanechnikov kernel, as suggested by Theorem~\ref{theorem:Theorem4}, was confirmed.

\section{Conclusion}\label{section:sec6}
First, we have found that a Biweight kernel is optimal in the sense of minimizing the AMSE of the MLR parameter estimator (Theorem~\ref{theorem:Theorem2}), 
via refining the theorem on the asymptotic normality of it (Theorem~\ref{theorem:Theorem1}).
Truncated kernels are generally more suitable for the MLR than heavy-tailed kernels (Table~\ref{tb:OPT}).

Secondly, we have analyzed the IRLS, which is applicable to a wider kernel class opposed to the MEM which gives an analytic parameter update only for a Gaussian kernel.
We have provided a sufficient condition under which the objective function sequence given via the IRLS converges (Theorem~\ref{theorem:Theorem3}).
Moreover, we have shown that both of the objective function and parameter estimate sequences obtained by the IRLS converge exactly in a finite number of iterations when an Epanechnikov kernel is used (Theorem~\ref{theorem:Theorem4}).

In the simulation, among those kernels investigated, the MSE under the large sample size became the smallest for a Biweight kernel and the computational efficiency of the IRLS was best for an Epanechnikov kernel, as suggested in our theorems.

From the above results, one would be able to recommend a Biweight kernel or an Epanechnikov kernel for use in the MLR; 
practitioners can select them properly according to their needs, on the basis of the findings in this paper.
Additionally in order to use them, it will be reasonable to proceed with discussion in the framework of the IRLS-based parameter estimation.

\section*{Appendix}
\subsection{Proof Outline of Theorem~\ref{theorem:Theorem1}}
\hypertarget{A}{The} Taylor expansion of $\nabla O(\bm{\theta})$ at $\hat{\bm{\theta}}_n$ around $\tilde{\bm{\theta}}$ is
\begin{align}
\label{eq:tylor}
\bm{0}
=\nabla O(\hat{\bm{\theta}}_n)
=\bm{b}_n+\mathrm{A}_n(\hat{\bm{\theta}}_n-\tilde{\bm{\theta}}),
\end{align}
where $\bm{b}_n=\nabla O(\tilde{\bm{\theta}})$, 
$\mathrm{A}_n=\nabla^2 O(\bm{\theta}^*)$, 
and where an appropriately determined $\bm{\theta}^*$ satisfying $\|\bm{\theta}^*-\tilde{\bm{\theta}}\|\le\|\hat{\bm{\theta}}_n-\tilde{\bm{\theta}}\|$.
The equation~\eqref{eq:tylor} implies $\hat{\bm{\theta}}_n-\tilde{\bm{\theta}}=-\mathrm{A}_n^{-1}\bm{b}_n$ if $\mathrm{A}_n$ is regular.

\cite{kemp2012regression} has proved that $\mathrm{A}_n$ is a consistent estimator of $\tilde{\mathrm{A}}$.
Although the asymptotic normality of $\bm{b}_n$ has been also proved, 
its mean reduces to $\mathbf{0}$ due to their strong assumption $h_n=o(n^{-1/7})$.
Thus, it is enough to clarify the mean of $\bm{b}_n$ and requirements for the proof, 
under the weaker assumptions. 
These can be proved on the basis of the Taylor expansion-based analysis framework.
The regularity conditions of Theorem~\ref{theorem:Theorem1}, which are same as ones given in \cite{kemp2012regression}, are
\begin{itemize}\setlength{\parskip}{0cm}\setlength{\itemsep}{0cm} 
\item $\{(\bm{x}_i,y_i)\in\mathbb{R}^p\times\mathbb{R}\}_{i=1}^n$ is an iid sequence.
\item $\mathrm{E}[\|\bm{x}\|^{5+\xi}]<\infty$ for some $\xi>0$.
\item $\pYX(y|\bm{x})$ is continuous in $y$ and $\pYX(y|\bm{x})\le\pYX(\tilde{\bm{\theta}}^\top\bm{x}|\bm{x})$ for all $y$ and $\bm{x}$. 
In addition, there exists a set $S\in\mathbb{R}^p$ such that ${\rm Pr}[\bm{x}\in S]=1$ and 
$\pYX(y|\bm{x})<\pYX(\tilde{\bm{\theta}}^\top\bm{x}|\bm{x})$ for all $y\neq\tilde{\bm{\theta}}^\top\bm{x}$ and $\bm{x}\in S$.
\item $\pYX^{(j)}(y|\bm{x})$, $j=0,\ldots,3$ are uniformly bounded.
\item $\tilde{\mathrm{A}}$ is negative definite.
\item The parameter space is a compact space that includes $\tilde{\bm{\theta}}$ in its interior.
\end{itemize}
Also, it should be noted that \cite{yao2014new} has shown a Gaussian case and that the same analysis method has appeared in mode estimation~\cite{eddy1980optimum, romano1988weak}, and these works are also a reference.

\subsection{Proof Outline of Theorem~\ref{theorem:Theorem2}}
\hypertarget{B}{Theorem}~\ref{theorem:Theorem2} can be proved by applying in order Theorem~1~(v), (iii), and Theorem~C in~\cite{granovsky1991optimizing}. 
The AMSE criterion $U^{6/7}V^{4/7}$ is invariant
of the scaling of the kernel function,
so that a Biweight kernel is optimal even if it is arbitrarily scaled. 
The compact support is not presumed.

\subsection{Proof of Theorem~\ref{theorem:Theorem4}}
\hypertarget{C}{Let} $N=\{1,\ldots,n\}$, and $\mathcal{I}(\bm{\theta})$ be the set of indices for which the argument $y_i-\bm{\theta}^{\top}\bm{x}_i$
of $K_h(y_i-\bm{\theta}^{\top}\bm{x}_i)$ is in the non-flat region
of the kernel $K_h$:
\begin{align}
\mathcal{I}(\bm{\theta})
=\left\{i\in N:\left|y_i-\bm{\theta}^{\top}\bm{x}_i\right|\le h\right\}\subset N.
\end{align}
Thus, the objective function~\eqref{eq:OBJ} is written as 
\begin{align}
O(\bm{\theta})
=\frac{c_h|\mathcal{I}(\bm{\theta})|}{n}
-\frac{c_h}{nh^2}\sum_{i\in\mathcal{I}(\bm{\theta})}\left(y_i-\bm{\theta}^{\top}\bm{x}_i\right)^2,
\end{align}
and the corresponding update of the IRLS~\eqref{eq:MM-IRLS} becomes
\begin{align}
\label{eq:Epa-iter}
\bm{\theta}_{t+1}
=\left(\sum_{i\in\mathcal{I}_t}\bm{x}_i\bm{x}_i^{\top}\right)^{-1}
\sum_{i\in\mathcal{I}_t}y_i\bm{x}_i,
\end{align}
where $c_h=3/(4h)$ and we let $\mathcal{I}_t$ to denote $\mathcal{I}(\bm{\theta}_t)$.
For any $\bm{\theta}$, $\mathcal{I}(\bm{\theta})$ is in $\mathcal{P}(N)$,
the power set of $N$.
We note that, given a sample set, $\bm{\theta}_{t+1}$ depends on
$\bm{\theta}_t$ only through $\mathcal{I}_t$.
Thus, at most $|\mathcal{P}(N)|=2^n$ different values
appear in the parameter estimate sequence $\{\bm{\theta}_t\}$
as well as the sequence $\{O(\bm{\theta}_t)\}$.
Convergence of $\{O(\bm{\theta}_t)\}$ is guaranteed by Theorem~\ref{theorem:Theorem3}.
Since a finite number of values appear in $\{O(\bm{\theta}_t)\}$, 
there exists $\tau$ such that for all $t\ge\tau$, $O(\bm{\theta}_t)$ is a constant. 
In the following, we prove that $\{\bm{\theta}_t\}$ also converges at $\tau$th iteration.
Introduce the best quadratic minorizer of $O(\bm{\theta})$ at $\bm{\theta}'$ as 
\begin{align}
O_M(\bm{\theta}|\bm{\theta}')
=\frac{c_h|\mathcal{I}(\bm{\theta}')|}{n}
-\frac{c_h}{nh^2}\sum_{i\in\mathcal{I}(\bm{\theta}')}\left(y_i-\bm{\theta}^{\top}\bm{x}_i\right)^2.
\end{align}
The inequality~\eqref{eq:Mino} implies 
\begin{align}
O(\bm{\theta}_{\tau})=O_M(\bm{\theta}_{\tau}|\bm{\theta}_{\tau});\quad
O(\bm{\theta}_{\tau+1}) \ge O_M(\bm{\theta}_{\tau+1}|\bm{\theta}_{\tau}),
\end{align}
and combining them leads 
\begin{align}
\label{eq:inq1}
O(\bm{\theta}_{\tau+1})-O(\bm{\theta}_{\tau})
\ge O_M(\bm{\theta}_{\tau+1}|\bm{\theta}_{\tau})
-O_M(\bm{\theta}_{\tau}|\bm{\theta}_{\tau})\ge0.
\end{align}
Since $O(\bm{\theta}_t)$ is constant for $t\ge\tau$,
$O_M(\bm{\theta}_{\tau+1}|\bm{\theta}_{\tau})-O_M(\bm{\theta}_{\tau}|\bm{\theta}_{\tau})$ should be equal to 0.
Alternatively, from~\eqref{eq:Epa-iter} one has 
\begin{align}
\begin{split}
&O_M(\bm{\theta}_{\tau+1}|\bm{\theta}_{\tau})-O_M(\bm{\theta}_{\tau}|\bm{\theta}_{\tau})\\
&=\frac{c_h}{nh^2}
\left(\bm{\theta}_{\tau+1}-\bm{\theta}_{\tau}\right)^{\top}
\left(\sum_{i\in\mathcal{I}_\tau}\bm{x}_i\bm{x}_i^\top\right)
\left(\bm{\theta}_{\tau+1}-\bm{\theta}_{\tau}\right).
\end{split}
\end{align}
Since the left-hand side equals to 0,
one has $\bm{\theta}_{\tau+1}=\bm{\theta}_\tau$
and consequently $\bm{\theta}_t$ is constant for $t\ge\tau$, 
under the additional condition that the Hessian $\propto-\sum_{i\in\mathcal{I}_\tau}\bm{x}_i\bm{x}_i^\top$ of $O(\bm{\theta})$ at $\bm{\theta}=\bm{\theta}_\tau$ is negative definite.
This completes the proof.
\hfill$\Box$

The Hessian of $O(\bm{\theta}_\tau)$ is written as $\mathrm{A}_n$ in the Appendix~\hyperlink{A}{A}.
Under the asymptotic situation satisfying the conditions of Theorem~\ref{theorem:Theorem1}, 
the Hessian $\mathrm{A}_n$ gets negative definite, 
and hence the latter half of Theorem~\ref{theorem:Theorem4} holds.
Also, this convergence result still holds even if using the update $\bm{\theta}_{t+1}=
(\mathrm{X}^{\top}\mathrm{G}_{h t}\mathrm{X}-\epsilon\mathrm{I})^{-1}\mathrm{X}^{\top}\mathrm{G}_{ht}\bm{y}$ with $\epsilon>0$ and an identity matrix $\mathrm{I}$.
This form corresponds to a $\ell_2$-regularized version of the MLR or a technique for stabilization of inverse matrix calculation.
In this case, the additional condition is not needed.

Finally, we would like to note that this way to prove has been similarly conducted for the analysis~\cite{comaniciu1999mean, Huang2018} of the mean shift algorithm (MS), 
which is used for mode estimation and so on.
The MS also can be viewed as a MM-based optimization of a kernel-based objective function.
Additionally, the relationship between the MS and the MEM for mode estimation is similar to that of the IRLS and the MEM for the MLR; 
the MS is more generic~\cite{yamasaki2019properties}.
The MEM transforms problems that are easier to directly optimize into harder problems.

\bibliography{article}

\begin{thebibliography}{10}
\providecommand{\url}[1]{#1}
\csname url@samestyle\endcsname
\providecommand{\newblock}{\relax}
\providecommand{\bibinfo}[2]{#2}
\providecommand{\BIBentrySTDinterwordspacing}{\spaceskip=0pt\relax}
\providecommand{\BIBentryALTinterwordstretchfactor}{4}
\providecommand{\BIBentryALTinterwordspacing}{\spaceskip=\fontdimen2\font plus
\BIBentryALTinterwordstretchfactor\fontdimen3\font minus
  \fontdimen4\font\relax}
\providecommand{\BIBforeignlanguage}[2]{{%
\expandafter\ifx\csname l@#1\endcsname\relax
\typeout{** WARNING: IEEEtran.bst: No hyphenation pattern has been}%
\typeout{** loaded for the language `#1'. Using the pattern for}%
\typeout{** the default language instead.}%
\else
\language=\csname l@#1\endcsname
\fi
#2}}
\providecommand{\BIBdecl}{\relax}
\BIBdecl

\bibitem{lee1989mode}
M.-J. Lee, ``Mode regression,'' \emph{Journal of Econometrics}, vol.~42, no.~3,
  pp. 337--349, 1989.

\bibitem{lee1993quadratic}
------, ``Quadratic mode regression,'' \emph{Journal of Econometrics}, vol.~57,
  no. 1--3, pp. 1--19, 1993.

\bibitem{kemp2012regression}
G.~C. Kemp and J.~{Santos Silva}, ``Regression towards the mode,''
  \emph{Journal of Econometrics}, vol. 170, no.~1, pp. 92--101, 2012.

\bibitem{yao2014new}
W.~Yao and L.~Li, ``A new regression model: modal linear regression,''
  \emph{Scandinavian Journal of Statistics}, vol.~41, no.~3, pp. 656--671,
  2014.

\bibitem{baldauf2012use}
M.~Baldauf and J.~{Santos Silva}, ``On the use of robust regression in
  econometrics,'' \emph{Economics Letters}, vol. 114, no.~1, pp. 124--127,
  2012.

\bibitem{NIPS20176743}
X.~Wang, H.~Chen, W.~Cai, D.~Shen, and H.~Huang, ``Regularized modal regression
  with applications in cognitive impairment prediction,'' in \emph{Advances in
  Neural Information Processing Systems}, 2017, pp. 1448--1458.

\bibitem{feng2017statistical}
Y.~Feng, J.~Fan, and J.~A. Suykens, ``A statistical learning approach to modal
  regression,'' \emph{arXiv preprint arXiv:1702.05960}, 2017.

\bibitem{tian2017fitting}
M.~Tian, J.~He, and K.~Yu, ``Fitting truncated mode regression model by
  simulated annealing,'' in \emph{Computational Optimization in
  Engineering-Paradigms and Applications}.\hskip 1em plus 0.5em minus
  0.4em\relax IntechOpen, 2017.

\bibitem{sando2018information}
K.~Sando, S.~Akaho, N.~Murata, and H.~Hino, ``Information geometric perspective
  of modal linear regression,'' in \emph{International Conference on Neural
  Information Processing}.\hskip 1em plus 0.5em minus 0.4em\relax Springer,
  2018, pp. 535--545.

\bibitem{ohta2018quantile}
H.~Ohta, K.~Kato, and S.~Hara, ``Quantile regression approach to conditional
  mode estimation,'' \emph{arXiv preprint arXiv:1811.05379}, 2018.

\bibitem{zhao2014robust}
W.~Zhao, R.~Zhang, J.~Liu, and Y.~Lv, ``Robust and efficient variable selection
  for semiparametric partially linear varying coefficient model based on modal
  regression,'' \emph{Annals of the Institute of Statistical Mathematics},
  vol.~66, no.~1, pp. 165--191, 2014.

\bibitem{Salah2017}
S.~Khardani and A.~F. Yao, ``Non linear parametric mode regression,''
  \emph{Communications in Statistics --- Theory and Methods}, vol.~46, no.~6,
  pp. 3006--3024, 2017.

\bibitem{LI201815}
X.~Li and D.~Zhu, ``Robust feature selection via $l_{2,1}$-norm in finite
  mixture of regression,'' \emph{Pattern Recognition Letters}, vol. 108, no.~1,
  pp. 15--22, 2018.

\bibitem{li2007nonparametric}
J.~Li, S.~Ray, and B.~G. Lindsay, ``A nonparametric statistical approach to
  clustering via mode identification,'' \emph{Journal of Machine Learning
  Research}, vol.~8, pp. 1687--1723, 2007.

\bibitem{yu2012bayesian}
K.~Yu and K.~Aristodemou, ``Bayesian mode regression,'' \emph{arXiv preprint
  arXiv:1208.0579}, 2012.

\bibitem{liu2007correntropy}
W.~Liu, P.~P. Pokharel, and J.~C. Pr{\'\i}ncipe, ``Correntropy: Properties and
  applications in non-gaussian signal processing,'' \emph{IEEE Transactions on
  Signal Processing}, vol.~55, no.~11, pp. 5286--5298, 2007.

\bibitem{feng2015learning}
Y.~Feng, X.~Huang, L.~Shi, Y.~Yang, and J.~A. Suykens, ``Learning with the
  maximum correntropy criterion induced losses for regression.'' \emph{Journal
  of Machine Learning Research}, vol.~16, pp. 993--1034, 2015.

\bibitem{granovsky1991optimizing}
B.~L. Granovsky and H.-G. M{\"u}ller, ``Optimizing kernel methods: a unifying
  variational principle,'' \emph{International Statistical Review/Revue
  Internationale de Statistique}, vol.~59, no.~3, pp. 373--388, 1991.

\bibitem{lange2016mm}
K.~Lange, \emph{MM Optimization Algorithms}.\hskip 1em plus 0.5em minus
  0.4em\relax SIAM, 2016.

\bibitem{de2009sharp}
J.~de~Leeuw and K.~Lange, ``Sharp quadratic majorization in one dimension,''
  \emph{Computational Statistics \& Data Analysis}, vol.~53, no.~7, pp.
  2471--2484, 2009.

\bibitem{eddy1980optimum}
W.~F. Eddy \emph{et~al.}, ``Optimum kernel estimators of the mode,'' \emph{The
  Annals of Statistics}, vol.~8, no.~4, pp. 870--882, 1980.

\bibitem{romano1988weak}
J.~P. Romano, ``On weak convergence and optimality of kernel density estimates
  of the mode,'' \emph{The Annals of Statistics}, vol.~16, no.~2, pp. 629--647,
  1988.

\bibitem{comaniciu1999mean}
D.~Comaniciu and P.~Meer, ``Mean shift analysis and applications,'' in
  \emph{IEEE International Conference on Computer Vision}, vol.~2.\hskip 1em
  plus 0.5em minus 0.4em\relax IEEE, 1999, pp. 1197--1203.

\bibitem{Huang2018}
K.~Huang, X.~Fu, and N.~Sidiropoulos, ``On convergence of {E}panechnikov mean
  shift,'' in \emph{AAAI Conference on Artificial Intelligence}, 2018.

\bibitem{yamasaki2019properties}
R.~Yamasaki and T.~Tanaka, ``Properties of mean shift,'' \emph{IEEE
  Transactions on Pattern Analysis and Machine Intelligence}, 2019, early
  access.

\end{thebibliography}
\end{document}